\documentclass[12pt]{article}
\pdfoutput=1
\usepackage{jcappub}
\usepackage{slashed}
\usepackage{amsmath}
\usepackage[frak=euler,scr=boondox,bb= pazo]{mathalfa}
\usepackage{cancel}

\def\be{\begin{equation}}
\def\ee{\end{equation}}
\def\bea{\begin{eqnarray}}
\def\eea{\end{eqnarray}}

\begin{document}

\title{Towards a theory of quantum gravity from neural networks} 

\author[1,2]{Vitaly Vanchurin}

\emailAdd{vitaly.vanchurin@gmail.com}

\date{\today}

\affiliation[1]{National Center for Biotechnology Information, NIH, Bethesda, Maryland 20894, USA}

\affiliation[2]{Duluth Institute for Advanced Study, Duluth, Minnesota, 55804, USA}

\date{\today}

\abstract{Neural network is a dynamical system described by two different types of degrees of freedom: fast-changing non-trainable variables (e.g. state of neurons) and slow-changing trainable variables (e.g. weights and biases). We show that the non-equilibrium dynamics of trainable variables can be described by the Madelung equations, if the number of neurons is fixed, and by the Schrodinger equation, if the learning system is capable of adjusting its own parameters such as the number of neurons, step size and mini-batch size. We argue that the Lorentz symmetries and curved space-time can emerge from the interplay between stochastic entropy production and entropy destruction due to learning. We show that the non-equilibrium dynamics of non-trainable variables can be described by the geodesic equation (in the emergent space-time) for localized states of neurons, and by the Einstein equations (with cosmological constant) for the entire network. We conclude that the quantum description of trainable variables and the gravitational description of non-trainable variables are dual in the sense that they provide alternative macroscopic descriptions of the same learning system, defined microscopically as a neural network.  } 

\maketitle

\section{Introduction}

Quantum mechanics is a well-defined mathematical framework that proved to be very successful for modeling a wide range of complex phenomena in high energy and condensed matter physics, but it fails to give any reasonable explanations for a phenomenon as simple as a measurement, i.e. the measurement problem. It is completely unclear what is actually happening with the wave-function during the measurement and what role (if any) observers play in the process. Unfortunately, none of the current interpretations of quantum mechanics provide a satisfactory answer to the above questions. In the Copenhagen interpretation it is simply postulated that during measurement the wave-function undergoes a sudden collapse. That is fine, but then one should view quantum mechanics as a phenomenological theory with its limits of validity. In the many-worlds interpretation the wave-function describes the state of the entire universe which evolves unitarily and nothing ever collapses \cite{Everett}. That is an opposite view where quantum mechanics is a fundamental theory, but it is not a very useful theory as it makes no probabilistic predictions that could be checked experimentally. In the recent years, the so-called emergent quantum mechanics is becoming more popular \cite{Adler, tHooft, Blasone,Grossing,Acosta,deCordoba, Caticha, entropic}, but what is usually missing is a microscopic description of the dynamics from which the complex wave-function and the Schrodinger equation could emerge. Moreover, if quantum mechanics does emerge from a statistical theory, for example, due to averaging over some hidden variables \cite{Bohm}, then the hidden variables must be non-local \cite{Bell}. In this paper we describe a microscopic theory of neural networks from which the quantum behavior does emerge (for the trainable variables) and the hidden variables (or the non-trainable variables) are non-local \cite{worldasnn, quantumness}. In fact, as we shall see, the very notion of locality is also an emergent phenomenon that arises from the learning dynamics of neural networks. 

General relativity is another well-defined mathematical framework that was developed for modeling a wide range of astrophysical and cosmological phenomena, but it is also incomplete since it does not describe what happens in space-time singularities and it does not directly explain the indirect observations of dark matter, dark energy and cosmic inflation. Of course, we can also treat general relativity as a (highly successful, but still) phenomenological theory with its own limits of validity and model all of these phenomena with phenomenological fields, but then certain important questions cannot be answered. And that includes not only very general questions about the nature of dark matter, dark energy and cosmic inflation, but also more specific questions about assigning probabilities to cosmological observations, i.e. the measure problem \cite{measure}. Perhaps, in a more complete theory of quantum gravity all of these questions would have answers and, in fact, some progress in developing such a theory had been made in context of AdS/CFT \cite{Maldacena, Witten, Susskind} and loop quantum gravity \cite{Ashtekar, Rovelli, ABL}. Another possibility is that gravity is an emergent phenomenon \cite{Jacobson, Verlinde, Padmanabhan, emergent} similar to thermodynamics, and then it does not make sense to quantize the metric tensor as all other fields, but, instead, we should try to figure out from which microscopic theory the general theory of relativity could emerge.  In this paper we describe how not only general relativity, but also quantum mechanics, Lorentz invariance and space-time can all emerge from the learning dynamics of neural networks \cite{worldasnn}. Note that the idea of using neural networks to describe gravity was also explored in Ref. \cite{Dvali} in the context of quantum neural networks and black holes, and in Ref. \cite{Smolin} in the context of matrix models and cosmology. 

The paper is organized as follows. In the following section we define the microscopic theory of neural networks and develop a statistical description of the learning phenomenon. In Sec. \ref{sec:madelung} we derive Madelung equations which can be used for modeling the dynamics of trainable variables both in and out of equilibrium. In Sec. \ref{sec:schrodinger} we show that if the learning system is capable of adjusting its own parameters such as step size, mini-batch size and/or acceptance of neurons, then the trainable variables must evolve according to the Schrodinger equation. In Sec. \ref{sec:lorentz} we consider the dynamics of non-trainable variables of individual neurons to show how the null, time-like and space-like vectors emerge. In Sec. \ref{sec:spacetime} we exploit the freedom of local transformations to define an emergent space-time and the metric tensor. In Sec. \ref{sec:geodesic} we consider minimally interacting states of neurons to show that the neurons must move along geodesics in the emergent space-time. In Sec. \ref{sec:einstein} we argue that the dynamics of non-trainable variables in the entire network must be described by an action which is equivalent to the Einstein-Hilbert action up to a boundary term and by the cosmological constant which imposes a constraint on the number of neurons.  The main results of the paper are discussed in Sec. \ref{sec:discussion}.

\section{Neural networks}\label{sec:learning}

In general, a neural network can be defined as a septuple $({\bf x}, \hat{P}, p_\partial, \hat{w}, {\bf b}, {\bf f}, H)$, where:
\begin{enumerate}
\item ${\bf x}$, column state vector of boundary (i.e. input/output) and bulk (i.e. hidden) neurons,  
\item $\hat{P}$, boundary projection operator to subspace spanned by boundary neurons,
\item $p_\partial\left(\hat{P} {\bf x} \right )$, probability distribution which describes the training dataset, 
\item $\hat{w}$, weight matrix which describes connections and interactions between neurons, 
\item ${\bf b}$, column bias vector which describes bias in the inputs of individual neurons, 
\item ${\bf f}({\bf y})$, activation map which describes a non-linear part of the dynamics,
\item ${H}({\bf x}, {\bf b}, \hat{w}) = {H}({\bf x}, {\bf q}) $, loss function where $ {\bf q}$ denotes collectively both ${\bf b}$ and $\hat{w}$.
\end{enumerate} 
(See Ref. \cite{learningtheory} for details.) There are two types of degrees of freedom: trainable variables ${\bf q}$ (or  the bias vector ${\bf b}$ and weight matrix $\hat{w}$) and non-trainable variables ${\bf x}$ (or the state of boundary $\hat{P} {\bf x}$ and bulk  $ (\hat{I} -  \hat{P}) {\bf x}$ neurons). The state of the boundary neurons is updated either periodically or randomly from a training dataset which is described by some probability distribution $p_\partial({\bf x}_\partial)$, and between the updates both bulk and boundary neurons evolve according to 
\be
{\bf x}({t}) = {\bf f} \left(\hat{w} {\bf x}(t-1)+ {\bf b} \right),\label{eq:bulk_eom}
\ee
where the activation map acts separately on each component, i.e. $f_i ( {\bf y} ) = f_i(y_i) $ (e.g. hyperbolic tangent $\tanh(y)$, rectifier linear unit function $\max(0,x)$, etc.)

The main objective of learning is to find the trainable variables ${\bf q}$ (or  the bias vector ${\bf b}$ and weight matrix $\hat{w}$) which minimize the time-average of some loss function. For example, the boundary loss function is 
\bea
H({\bf x}, {\bf q}) &=&  \frac{1}{2} \left ( {\bf x}  - {\bf f} \left ( \hat{w} {\bf x}+ {\bf b} \right) \right )^T  \hat{P} \left (  {\bf x}  - {\bf f} \left ( \hat{w} {\bf x}+ {\bf b} \right) \right ) 
\label{eq:boundry_loss} 
\eea
and the bulk loss function can be defined as
\bea
H({\bf x}, {\bf q}) &=&  \frac{1}{2} \left ( {\bf x}  - {\bf f} \left ( \hat{w} {\bf x}+ {\bf b} \right) \right )^T \left (  {\bf x}  - {\bf f} \left ( \hat{w} {\bf x}+ {\bf b} \right) \right ) + \frac{1}{2} {V}({\bf x}, {\bf q}) 
\label{eq:bulk_loss} 
\eea
where in addition to the first term, which represents a sum of local errors over all neurons, there may be a second term which represents either local objectives or constraints imposed by a neural architecture  \cite{learningtheory}. Note that the boundary loss is usually used in supervised learning, but the bulk loss may be used for both supervised and unsupervised learning tasks.

To develop a statistical description of learning \cite{learningtheory}, consider a joint probability distribution over both trainable ${\bf q}$ and non-trainable ${\bf x}$ variables,
\be
p({\bf x}, {\bf q}) = p({\bf q}) p({\bf x}|{\bf q}),
\ee
where $p({\bf q})$ and $p({\bf x}|{\bf q})$ denote, respectively, the marginal and conditional distributions. If the non-trainable variables quickly equilibrate, then their distribution must be given by the maximum entropy distribution \cite{maxent}, 
\be
p({\bf x}|{\bf q}) \propto p_\partial({\bf x})  \exp\left (-\beta H({\bf x}, {\bf q}) \right),
\ee
where $\beta$ is a Lagrange multiplier which imposes a constraint on the loss function, 
\be
U({\bf q}) = \int d^N x \;p({\bf x}|{\bf q})  H({\bf x}, {\bf q}).
\ee
The corresponding free energy is
\be
F({\bf q},t) = \frac{1}{\beta} \log \left (\int d^N x \;p_\partial({\bf x},t)  \exp\left (-\beta H({\bf x}, {\bf q}) \right) \right ),
\ee
where the explicit and implicit dependencies of the free energy $F({\bf q}, t)$ on time $t$ are due to, respectively, stochastic dynamics of $p_\partial({\bf x},t)$ and learning dynamics of ${\bf q}(t)$. The total change of the free energy is given by 
\bea
\frac{d}{dt} F({\bf q}, t) &=& \sum_k \frac{d q_k}{d t} \frac{\partial F({\bf q}, t)}{\partial q_k}+ \frac{\partial F({\bf q}, t)}{\partial t}  \notag\\
 &=&  - \gamma \sum_k \left ( \frac{\partial F({\bf q}, t)}{\partial q_k} \right )^2   +\frac{\partial F({\bf q}, t)}{\partial t},  \label{eq:dF}
\eea
where we assume that the trainable variables experience a classical drift in the direction of the gradient of the free energy,
\be
 \frac{d q_{k}}{d t}  =  - \gamma \frac{\partial F({\bf q}, t)}{\partial q_k}  \label{eq:dq}.
\ee
Note that the parameter $\gamma$ can be either positive or negative depending on whether the free energy is minimized or maximized with resect to a given trainable variable. If evolution is dominated by stochastic dynamics, then according to the second law of thermodynamics the entropy must increase and then the free energy is minimized, but if evolution is dominated by learning, then according to the second law of learning the entropy must decrease and then the free energy can be maximized \cite{learningtheory}. We will come back to the issue of sign of $\gamma$ in the following sections.

\section{Madelung equations}\label{sec:madelung}

On the shortest time scales (or when the free energy $F({\bf q}, t)$ does not change significantly) the dynamics of the probability density $p({\bf q}, t)$ can be modeled by the Fokker-Planck equation,
\bea
\frac{\partial p({\bf q}, t)}{\partial t}&=&  \sum_k \frac{\partial}{\partial q_{k}}  \left (D  \frac{\partial p({\bf q}, t)}{\partial q_{k}} - \frac{d {q}_k}{dt} p({\bf q}, t) \right)\notag\\
&=&  \sum_k \frac{\partial}{\partial q_{k}}  \left (D  \frac{\partial p({\bf q}, t)}{\partial q_{k}} + \gamma \frac{\partial F({\bf q}, t)}{\partial q_k}  p({\bf q}, t) \right),
\label{eq:dP}
\eea
where we used \eqref{eq:dq} and $D$ is the diffusion coefficient. On longer time scales the dynamics of the free energy is given by \eqref{eq:dF} and an additional assumption must be made. By following the analysis of Refs. \cite{learningtheory, worldasnn, quantumness} we assume that the long time scale dynamics is governed by the principle of stationary entropy production \cite{entropic}. The principle states that the path taken by a system is the one for which the entropy production is stationary, subject to whatever constraints are imposed on the system.

The entropy production of trainable variables ${\bf q}$ can be estimated by calculating the total change in the Shannon entropy,
\be
S_q \equiv - \int d^K q\; p({\bf q}, t) \log p({\bf q}, t),
\ee
which can be expressed as
\bea
\Delta S_q &=&  \int_0^T dt \frac{d S_q(t) }{d t}  = - \int_0^T dt\, d^K q   \frac{d p}{dt}  \log p  - \int_0^T dt\, d^K q   \frac{d p}{dt}  \\
&=& \int_0^T dt\, d^K q  \, p  \left ( D \sum_k \left (\frac{\partial \log p}{\partial q_{k}}  \right )^2 + \gamma \sum_k   \frac{\partial \log p}{\partial q_{k}} \frac{\partial F}{\partial q_k} \right ). \label{eq:short}
\eea
where in the first line we used conservation of probabilities, i.e. $\int d^Kq\; p({\bf q}, t)= 1$ and in the second line we used \eqref{eq:dP}, integrated by parts and neglected the boundary terms by imposing either periodic or vanishing boundary conditions. Equation \eqref{eq:short} describes the system on short time scales, but on longer time scales an addition constraint must be imposed to satisfy \eqref{eq:dF}. The overall optimization problem is solved by constraining deviations of the free energy production \eqref{eq:dF} from its time-averaged value using the method of Lagrange multipliers. The corresponding ``action'' functional is given by,
\bea
{{\cal S}}[p, F, \alpha] &=&   \Delta S_q - \alpha\int d^K q  \;p \left (   \int_0^T dt \left ( \frac{dF}{d t}   -  \left \langle \frac{dF}{dt} \right \rangle_t \right )  - f  \right ),\notag\\
&=&   \int_0^T dt\, d^K q  \, p  \left ( D \sum_k \left (\frac{\partial \log p}{\partial q_{k}}  \right )^2 + \gamma \sum_k   \frac{\partial \log p}{\partial q_{k}} \frac{\partial F}{\partial q_k} + \alpha \gamma \sum_k \left ( \frac{\partial F}{\partial q_k} \right )^2  - \alpha \frac{\partial F}{\partial t}   + \alpha  {V} \right) \notag\\
&=&   \int_0^T dt\, d^K q  \, p  \left ( \left ( D -\frac{ \gamma}{4\alpha} \right )\sum_k \left (\frac{\partial \log p}{\partial q_{k}}  \right )^2 + \alpha \gamma \sum_k \left ( \frac{\partial \tilde{F}}{\partial q_k} \right )^2  - \alpha \frac{\partial \tilde{F}}{\partial t}   + \alpha  {V} \right). \label{eq:action}
\eea
In the second line we defined the ``potential''
\be
{V({\bf q})} \equiv  f  +  \left \langle \frac{dF({\bf q}, t)}{dt}  \right \rangle_t, \label{eq:potential}
 \ee
where $\left \langle ... \right \rangle_t $ is the time average, and in the third line we completed the square to define
 \be
 \tilde{F}({\bf q}, t) \equiv F({\bf q}, t) + \frac{\log p({\bf q}, t) }{2 \alpha}\label{eq:tilde_free}
 \ee
 and also used conservation of probabilities, i.e. $\int d^Kq\; p({\bf q}, t)= 1$. 
 
 Note that in Refs. \cite{learningtheory, worldasnn, quantumness} a functional similar to  \eqref{eq:action} was obtained, but only in a near equilibrium limit or when the entropy production due to learning is negligible, i.e. $\left | \gamma \sum_k \frac{\partial^2 F}{\partial q_k^2} \right | \ll D \sum_k \left ( \frac{\partial \log p}{\partial q_k} \right )^2$. In contrast, in \eqref{eq:action} we completed the square and redefined the free energy $F \rightarrow \tilde{F}$ which allowed us to keep all the terms. By varying  ${{\cal S}}[p, \tilde{F}, \alpha]$ with respect to $p$ and $\tilde{F}$ (i.e. original probability distribution, but shifted free energy \eqref{eq:tilde_free}) we also obtain the Madelung hydrodynamic equations \cite{Madelung}, 
 \bea
\frac{\partial}{\partial t} p &=& - \sum_k \frac{\partial}{\partial q_k} \left ( u_k p \right )  \label{eq:neural_FP}\\
\frac{\partial}{\partial t} u_j &=& -   \sum_k u_k \frac{\partial}{\partial q_k} u_j  - \frac{1}{M} \frac{\partial}{\partial q_j} \left ( V  - \frac{\hbar^2}{2 M}  \sum_k \frac{ \partial^2 \sqrt{p}}{\partial q_{k}^2} \right )\label{eq:neural_NS}
\eea
with ``velocity'' of the fluid
\be
u_k \equiv \frac{1}{M} \frac{\partial \tilde{F}}{\partial q_k} \label{eq:u}
\ee
and ``mass'' 
\be
M \equiv \frac{1}{2  \gamma},
\ee
but the ``Planck's constant'' is now
\be
 \hbar \equiv \frac{1}{\alpha} \sqrt{\frac{4 D}{ \gamma} \alpha -1} \label{eq:hbar}.
\ee
Therefore, we conclude, that the Madelung description of trainable variables must remain valid arbitrary far away from the learning equilibrium, suggesting that the effect is more general than previously thought. 
 
\section{Schrodinger equation}  \label{sec:schrodinger}
  
 All of the solutions of the Madelung equations \eqref{eq:neural_FP} and \eqref{eq:neural_NS} are also solutions of the Schrodinger equation, but the opposite is not true \cite{wallstrom} and so the system is not exactly quantum.  To study a limit when a fully quantum behavior emerges we have to assume that the learning system is described by a grand canonical ensemble of neurons and that the exact number of neurons $N$ is unobservable \cite{quantumness}. Then a constant shift in the free energy is unobservable, i.e.
\be
\tilde{F}  \cong \tilde{F} + \mu {N} \;\;\;\;\;\;\; \;\;\;\;\;\;\;\forall {N} \in \mathbb{Z}.\label{eq:first_cond}
\ee
where the ``chemical potential'' of neurons (or another Lagrange multiplier which imposes a constraint on the number of neurons) is defined as
\be
\mu =\pm {2\pi \hbar} \label{eq:chemical}.
\ee
Using \eqref{eq:first_cond}  the functional \eqref{eq:action} can be rewritten as,
\bea
{{\cal S}}[\Psi] =   {\alpha} \int_0^T dt\, d^K q \; \left ( \frac{\hbar^2}{2 M}   \sum_k\frac{\partial\Psi^*}{\partial q_k} \frac{\partial \Psi}{\partial q_k}   -  i  \hbar  \Psi^* \frac{ \partial \Psi }{\partial t}  +  {V} \Psi^*  \Psi  \right ), \label{eq:action4}
\eea
where the ``wave-function'' is
\be
\Psi \equiv \sqrt{p} \exp \left (- \frac{i \tilde{F}}{\hbar} \right ). \label{eq:wavefunction}
\ee
(See Ref.  \cite{quantumness} for details.) 

It is assumed that $\hbar$, $D$ and $\alpha$ are all positive, but $\gamma$ and $\mu$ can be either positive or negative.  By combining \eqref{eq:hbar} and \eqref{eq:chemical} we obtain a quadratic equation,
\be
\left (\frac{\mu}{2\pi} \right )^2 \alpha^2 -  \frac{4D}{\gamma} \alpha + 1 =0,
\ee
whose solutions are
\be
\alpha_{\pm} =  \frac{ \left ( \frac{2D}{\gamma} \right ) \pm \sqrt{ \left ( \frac{2 D}{ \gamma} \right )^2 - \left (\frac{\mu}{2\pi} \right )^2}  }{ \left (\frac{\mu}{2\pi} \right )^2}.
\ee
For the real solutions to exist, the following inequality must be satisfied
\be
 \left |  \frac{2D}{\gamma} \right | \ge \left | \frac{\mu}{2\pi} \right | = {\hbar} \label{eq:ineq}.
\ee
Evidently, for $\left | \gamma \mu / D \right | > 4\pi$ the inequality \eqref{eq:ineq} cannot be satisfied and thus the quantum (or Schrodinger, but not Madelung) description breaks down. To restore the quantumness the learning system must readjust $\gamma$, $\mu$ and/or $D$ such that the inequality  \eqref{eq:ineq}  is saturated. In other words the learning system must decrease either the step size, the mini-batch size and/or the chemical potential by $\gamma$, $D$ and $\mu$ until 
\be
\left |\frac{ \gamma \mu}{D} \right | = 4\pi . \label{eq:relation}
\ee
However, if we want the ``Planck constant'' to remain constant, then the chemical potential \eqref{eq:chemical} must be constant, and the only parameters that should vary are the number of neurons, the step size and the mini-batch size, or $N$, $\gamma$ and $D$. Evidently, the learning efficiently which is achievable only through quantumness (e.g. quantum annealing) is tightly connected to the ability of the learning system to dynamically adjust its own parameters (e.g. step size, mini-batch size, number of neurons). On the other hand the Madelung description is always appropriate both in and out of equilibrium. 

\section{Lorentz symmetry}\label{sec:lorentz}

In the previous sections we discussed the entropy production $\Delta S_q$ of trainable variables ${\bf q}$, but the dynamics of non-trainable variables was described only at the level of its free energy. In this section we are interested instead in the entropy production $\Delta S_x$ of non-trainable variables ${\bf x}$ which we approximate as a sum of entropy productions of individual neurons,
\be
\Delta S_x(t)\approx \sum_i \Delta S_{x,i}(t).
\ee
It is assumed that the state of neurons changes quasi-periodically \cite{worldasnn}, i.e. 
\be
{x}^1_i(t) \rightarrow {x}^2_i(t+1) \rightarrow ... \rightarrow {x}^d_i(t+d-1)\rightarrow x^1_i(t+d) ...\,.
\ee
For concreteness, we assume that $d=3$ which corresponds to the three spatial dimensions. Then, the entropy production can be modeled as a function of ``displacements'',
\be
\dot{x}_i^a(t) \equiv  \frac{1}{3} \left ( x_i^a(t+3) - x^a_i(t)\right),
\ee
and computational time $t$, i.e.
\be
\Delta S_{x,i}(t) =  \Delta S_{x,i}\left (\dot{x}^a_i(t), t \right ).
\ee
In general, there are two contributions to the entropy production: positive due to the second law of thermodynamics and negative due to the second law of learning \cite{learningtheory}. The main idea is to model the positive entropy production as some non-negative function $\sigma_{i,+}(t) \ge 0$ of computational time $t$ and the negative entropy production (or entropy destruction) as some non-positive function ${\sigma}_{i,-}(\dot{x}^a_i(t)) \le 0$ of displacements $\dot{x}^a_i(t)$.  Then the total entropy production is given by
\be
 \Delta S_{x,i}\left (\dot{x}^a_i(t), t \right ) = \sigma_{i,+}(t) + \sigma_{i,-}(\dot{x}^a_i(t)) =  \dot{x}_i^0(t)^2 + \sigma_{i,-}(\dot{x}^a_i(t)) \label{eq:dS10}
 \ee 
where we defined a monotonic function 
\be
{x}_i^0(t) \equiv \int^t_0 d\tau \sqrt{\sigma_{i,+}(\tau)}.
\ee
In addition, the entropy destruction $\sigma_{i,-}(\dot{x}^a_i(t))\le 0$ must vanish if there are no displacements $\dot{x}^a_i = (0,0,0)$ which implies that there are no zeroth and first order terms in a perturbative expansion around origin, i.e.
\be
\sigma_{i,-}(\dot{x}^a_i(t)) \approx - \sum_{a,b} g_{i, ab} \,\dot{x}_i^a(t) \dot{x}_i^b(t). \label{eq:dS11}
\ee
Here ${g}_{i,ab}(t)$ is some positive definite matrix and the displacements $\dot{x}_i^a(t) $ are assumed to be small so that the third order terms can be neglected. By substituting \eqref{eq:dS11} in \eqref{eq:dS10} we obtain
\be
 \Delta S_{x,i}\left (\dot{x}^a_i(t), t \right ) \approx \dot{x}_i^0(t)^2 - \sum_{a,b} g_{i, ab} \,\dot{x}_i^a(t) \dot{x}_i^b(t). \label{eq:dS} 
\ee 
and if we define temporal components of the matrix, $g_{i,00} =-1$ and $g_{i,0a} =g_{i,a0} =0$, then \eqref{eq:dS} takes a more covariant form\footnote{For brevity of notations summation over repeated raised and lowered indices is implied everywhere unless explicitly stated otherwise.}
\be
 \Delta S_{x,i}\left (\dot{x}^\mu_i(t) \right ) \approx - g_{i, \mu\nu} \,\dot{x}_i^\mu(t) \dot{x}_i^\nu(t). \label{eq:dS1} 
\ee 
for metric signature $(-,+,+,+)$. Note that for very large displacements the third order terms may become important and then the approximation in \eqref{eq:dS11} would break down and the Lorentz symmetry in \eqref{eq:dS1} would be broken.

It is convenient to think of $\dot{x}^{\mu}_i$ as a four-vector in the tangent space at the ``position'' of $i$'th neuron. Indeed, if, macroscopically, one can only observe the entropy (or entropy production), then we have Lorentz invariance in a sense that different representation of the four-vectors $\dot{x}^\mu_i$, that are connected to each other through Lorentz transformations, are indistinguishable. In a local equilibrium,  the stochastic entropy production $\dot{x}_i^0(t)^2$ is balanced by the entropy destruction due to learning $g_{i, ab} \,\dot{x}_i^a(t) \dot{x}_i^b(t)$ and the entropy remains constant. Therefore, the null displacement vectors, $g_{i, \mu\nu} \,\dot{x}_i^\mu(t) \dot{x}_i^\nu(t)=0$, describe neurons in equilibrium, $\Delta S_{x,i}=0$. Moreover, the time-like displacement vectors, $g_{i, \mu\nu} \,\dot{x}_i^\mu(t) \dot{x}_i^\nu(t)<0$, describe neurons for which stochastic dynamics dominates, $\Delta S_{x,i}>0$, and the space-like displacement vectors, $g_{i, \mu\nu} \,\dot{x}_i^\mu(t) \dot{x}_i^\nu(t)>0$, (if such displacement vectors can be stable) describe neurons for which learning dynamics dominates, $\Delta S_{x,i}<0$. 

\section{Emergent space-time}\label{sec:spacetime}

The local space-time coordinates of individual neurons, $x_i^{\mu'}$, can be transformed using shifts, rotations and boosts, i.e.
\be
x^\mu_i(t) = \Lambda_{i, \phantom{\mu}\mu'}^{\phantom{i,} \mu} {x}^{\mu'}_i(t) + a^{\mu}_i, \label{eq:trans}
\ee
where $\Lambda_{i, \phantom{\mu}\mu'}^{\phantom{i,} \mu}$ is a Lorentz matrix. If the matrix $g_{i, \mu\nu}$ is transformed using inverse Lorentz matrix $\Lambda^{\phantom{i,} \phantom{\mu}\mu'}_{i, \mu}$, 
\be
  g_{i, \mu\nu}  =  \Lambda^{\phantom{i,} \phantom{\mu}\mu'}_{i, \mu} \Lambda^{\phantom{i,} \phantom{\nu}\nu'}_{i, \nu} g_{i,\mu'\nu'},  \label{eq:trans2}
\ee
then the entropy production does not change, 
\bea
 \Delta S_{x,i} &=&  - g_{i, \mu\nu} \,\dot{x}_i^{\mu}(t) \dot{x}_i^{\nu}(t) \\
 &=&   - \left ( \Lambda^{\phantom{i,} \phantom{\mu}\mu'}_{i, \mu} \Lambda^{\phantom{i,} \phantom{\nu}\nu'}_{i, \nu} g_{i,\mu'\nu'}  \right ) \left (  \Lambda_{i, \phantom{\mu}\alpha'}^{\phantom{i,} \mu} \dot{x}^{\alpha'}_i(t) \right )\left (  \Lambda_{i, \phantom{\mu}\beta'}^{\phantom{i,} \mu} \dot{x}^{\beta'}_i(t)  \right ) \notag\\
 &=&  - g_{i, \mu'\nu'} \,\dot{x}_i^{\mu'}(t) \dot{x}_i^{\nu'}(t) .
\eea
(Note that we adopted a standard notation of primed-unprimed indices often used for coordinate transformations \cite{Carroll}).
The main idea is to exploit the freedom of transformations to make an appropriately weighted average of $g_{i,\mu\nu}$ matrices as close to the flat metric  $\eta_{\mu\nu}$ as possible, 
 \be
g_{\mu\nu}(t, x^1, x^2, x^3)  \equiv \frac{\sum_i g_{i,\mu\nu} \sqrt{g_i} e^{-  \frac{1}{2}  \left ( x^a - {x}^{a}_i(t) \right )  g_{i,ab} \left ( x^b - {x}^{b}_i(t) \right ) }}{\sum_i \sqrt{g_i} e^{-   \frac{1}{2} \left ( x^a - {x}^{a}_i(t) \right )  g_{i,ab} \left ( x^b - {x}^{b}_i(t) \right ) }} \sim \eta_{\mu\nu},\label{eq:metric}
 \ee
 where $g_i = \det \left (g_{i,ab} \right )$ and summation in the exponent is taken over only spatial components, $a,b = 1,2,3$. For simplicity, we assume that all of the local space-times are transformed into ``synchronous gauge'' with  global time coordinate 
\be
t = x^0 = x_i^0
\ee
and
\bea
g_{00}(x) &=& -1\notag\\
g_{a0}(x)&=&0 \\
g_{0a}(x) &=& 0.\notag
\eea
Note that from now on the coordinate time is denoted by $t = x^0$ which need not be the same as computational time. 

It is convenient to introduce the curly brackets notation,
 \be
 \left \{ ... \right \} \equiv \sum_i ... (2\pi)^{-3/2}e^{- \frac{1}{2} \left ( x^a - {x}^{a}_i(t) \right )  g_{i,ab} \left ( x^b - {x}^{b}_i(t) \right ) },\label{eq:curly}
 \ee
and then the (weighted average)  metric tensor \eqref{eq:metric} can be expressed as
\be
g_{\mu\nu}(x) = \frac{\left \{ g_{i, \mu\nu} \sqrt{g_i} \right \} }{\left \{\sqrt{g_i} \right \} }\label{eq:g}
\ee
and (weighted average) inverse metric tensor is defined as
 \bea
g^{\mu\nu}(x) &\equiv& \frac{\left \{ g^{\mu\nu}_i \sqrt{g_i} \right \} }{\left \{\sqrt{g_i} \right \} }. \label{eq:ginv}
\eea
It is not immediately clear what is the relation between $g^{\mu\nu}(x)$ and $g_{\mu\nu}(x)$, but if the emergent space-time is nearly flat \eqref{eq:metric}, then we can expand both \eqref{eq:ginv} and \eqref{eq:g} around flat metric to obtain
\bea
g_{i,\mu\nu} &=& \eta_{\mu\nu} + \epsilon h_{i,\mu\nu}  + {\cal O}(\epsilon^2)\label{eq:pert1}\\ 
g_{i}^{\mu\nu} &=& \eta^{\mu\nu} - \epsilon  \eta^{\mu\alpha}  \eta^{\nu\beta} h_{i,\alpha\beta} + {\cal O}(\epsilon^2) \label{eq:pert2}
\eea 
and verify that the product of the (weighted average) metric tensor and of the (weighted average) inverse metric tensor is indeed identity,
\bea
g^{\mu\nu} g_{\nu\lambda} &=& \frac{\left \{ g^{\mu\nu}_i \sqrt{g_i} \right \} }{\left \{\sqrt{g_i} \right \} } \frac{\left \{ g_{i, \nu\lambda} \sqrt{g_i} \right \} }{\left \{\sqrt{g_i} \right \} } \notag \\
& \approx&  \left (  \eta^{\mu\nu} - \epsilon \frac{  \{ \eta^{\mu\alpha}  \eta^{\nu\beta} h_{i,\alpha\beta} \sqrt{g_i}  \}}{\left \{\sqrt{g_i} \right \} }  \right ) \left (  \eta_{ \nu\lambda} + \epsilon \frac{  \{ h_{i, \nu\lambda}  \sqrt{g_i}  \}}{\left \{\sqrt{g_i} \right \} }  \right )  + {\cal O}(\epsilon^2)  \\
&=& \delta^\mu_\lambda   + {\cal O}(\epsilon^2). \notag
\eea

In general, the curly brackets \eqref{eq:curly} can be used for mapping discrete indices $i$ to continuous spatial coordinates $(x^1,x^2,x^3)$.  For example, the total number of neurons can be expressed as 
 \bea
\int d^D x \left \{ \sqrt{g_i} \right \}  = \sum_i  \int d^3 x \sqrt{\frac{g_i}{(2\pi)^3}} \;e^{-  \frac{1}{2} \left ( x^a - {x}^{a}_i(t) \right )  g_{i,ab} \left ( x^b - {x}^{b}_i(t) \right ) }  = \sum_i 1 = N,\label{eq:N} 
\eea
which suggests that 
 \bea
\sqrt{-g(x)}\;  &\equiv& \left \{ \sqrt{g_i} \right \} \label{eq:detg} 
\eea
should be interpreted as the number density of neurons in the emergent space. Moreover, using the perturbative expansions \eqref{eq:pert1} and \eqref{eq:pert2} we can check that the determinant of the metric tensor $g_{ab}$ is the same as the weighted sum of determinants of $g_{i,ab}$, i.e.
\bea
\sqrt{ - \det \left ( g_{\mu\nu}(x) \right )}  &=& \sqrt{ \det \left (g_{ab} \right )} \notag\\
&=& 1 + \frac{\epsilon}{2}  \text{Tr} h_{ab} + {\cal O}(\epsilon^2) \notag\\
 &=& \frac{\left \{\left ( 1+  \frac{\epsilon}{2}   \text{Tr}( h_{i, ab} )  \right )\sqrt{g_i} \right \} }{\left \{\sqrt{g_i} \right \} }+ {\cal O}(\epsilon^2)\notag\\
 &=& \frac{ \{ 1 \} +  \epsilon  \left \{ \text{Tr}( h_{i, ab}) \right \} }{ \{ 1 \} +  \frac{\epsilon}{2}   \left \{ \text{Tr} ( h_{i, ab} ) \right \}}+ {\cal O}(\epsilon^2)\\
 &= &  \{ \sqrt{ g_i} \} + {\cal O}(\epsilon^2) \notag \\
 &=& \sqrt{- g(x)}+ {\cal O}(\epsilon^2). \notag
 \eea

\section{Geodesic equation}\label{sec:geodesic}

The proper time of a given neuron can be identified with the square root of the entropy production  \eqref{eq:dS2}, i.e.
\bea
\Delta \tau(t) & = &  \sqrt{ \Delta S_{i,x} } \notag\\
 &\approx&  \sqrt{-g_{i, {\mu}{\nu}}  \,\dot{x}_i^\mu(t) \dot{x}_i^\nu(t)}.\label{eq:proper1}
 \eea
If we are interested in a more macroscopic and localized distribution of neurons, then their average entropy production can be approximated using the metric tensor \eqref{eq:g},
 \bea
\Delta \tau(t) &= &   \sqrt{ \langle \Delta S_{i,x}  \rangle } \notag\\
&\approx& \sqrt{-g_{{\mu}{\nu}}(x_i)  \,\dot{x}_i^\mu(t) \dot{x}_i^\nu(t)}.\label{eq:proper0}
\eea
In a continuum limit \eqref{eq:proper0} becomes,
\be
 \frac{d\tau}{dt} \approx \Delta \tau  = \sqrt{-g_{{\mu}{\nu}}(x_i)  \,\dot{x}_i^\mu(t) \dot{x}_i^\nu(t)},
\ee
which is usually expressed as a square of infinitesimal line element,
\be
d\tau^2 = -g_{{\mu}{\nu}}(x_i)  \,d {x}_i^\mu(t) d {x}_i^\nu(t).
\ee
By integrating the proper time from initial position $x^\mu_i(0)$ at time $t=0$ to final position $x^\mu_i(T)$ at time $t=T$ we obtain,
\bea
\tau[x^\mu_i(T) |\, x^\mu_i(0)] & \approx & \int_{0}^T dt  \frac{d\tau}{dt} \notag\\
&\approx& \int_0^T dt \sqrt{-g_{{\mu}{\nu}}(x_i)  \,\dot{x}_i^\mu(t) \dot{x}_i^\nu(t)}.\label{eq:proper2}
\eea
According to the principle of stationary entropy production, it is expected that the neuron would ``travel'' along a path (from initial $x^\mu_i(0)$ to final $x^\mu_i(T)$ position) which extremizes (in this case maximizes) the entropy production or, equivalently, the proper time $\tau[x^\mu_i(T) |\, x^\mu_i(0)]$.

By setting variations of the proper time \eqref{eq:proper2} with resect to the trajectory $x_i^{\mu}(t)$ to zero,
\be
\frac{\delta \tau[x^\mu_i(T) |\, x^\mu_i(0)] }{\delta x_i^{\mu}(t)} = 0
\ee
we obtain the geodesic equation 
\be
\frac{d^2 x_i^{\mu}}{dt^2} =  \left ( \Gamma^0_{\alpha\beta} \frac{d x_i^{\mu}}{dt} -  \Gamma^\mu_{\alpha\beta} \right )\frac{d x_i^{\alpha}}{dt} \frac{d x_i^{\beta}}{dt} \label{eq:geo}
\ee
or, equivalently, in terms of proper time
\be
\frac{d^2 x_i^{\mu}}{d\tau^2} = - \Gamma^\mu_{\alpha\beta} \frac{d x_i^{\alpha}}{d\tau} \frac{d x_i^{\beta}}{d\tau} 
\ee
where the Christoffel symbol is defined as
\be
\Gamma^\mu_{\alpha\beta} \equiv \frac{1}{2} g^{\mu\nu} \left (\frac{\partial g_{\nu\alpha}}{\partial x^\beta} +\frac{\partial g_{\nu\beta}}{\partial x^\alpha} - \frac{\partial g_{\alpha\beta}}{\partial x^\nu} \right ).
\ee
(See Ref. \cite{Carroll} for a detailed derivation of the geodesic equation.) This result suggests that in the limit of minimal interactions, described by the metric tensor $g_{{\mu}{\nu}}(x_i)$, the localized states of neurons are expected to move along geodesics in the emergent space-time. 

\section{Einstein equations}\label{sec:einstein}

In this section, we are interested in the total entropy production of the non-trainable variables in the entire neural network during global time interval $T$, 
\be
\Delta S_x[g]  \approx \int_0^T dt \; \sum_i \Delta S_{x, i}\left(\dot{x}_i(t), t\right)+  \int_0^T dt\;  \Delta S_{int} . \label{eq:total}
\ee
where $\Delta S_{int}$ the entropy production due to interactions between neurons. The entropy production of individual neurons  \eqref{eq:dS1} in the synchronous gauge is 
\be
 \Delta S_{x,i}\left (\dot{x}^\mu_i(t) \right ) \approx  1 -  g_{i, ab} \,\dot{x}_i^a(t) \dot{x}_i^b(t) ,\label{eq:dS2} 
\ee 
and then the total entropy production can be expressed as
\bea
\Delta S_x[g]  &\approx&  \int_0^T dt \; \sum_i \left ( 1 -  g_{i, ab} \frac{d x^a_i(t)}{dt}  \frac{d x^b_i(t)}{dt} \right )+  \int_0^T dt\; \Delta S_{int} \notag\\
 &=&   \sum_i   \int_0^T dt \; g_{i, ab} \frac{d^2 x^a_i(t)}{dt^2}  x^b_i(t)  + NT +  \int_0^T dt \;\Delta S_{int}
 \eea
where we integrated by parts and neglected the boundary term. We can also drop the constant term $NT$ (which is irrelevant for variational problems) and rewrite the entropy production using Gaussian integration formula and the definition of curly brackets \eqref{eq:curly},
  \bea
\Delta S_x[g]   &=&  \int_0^T  dt d^3 x  \left \{ g_{i, ab}  \frac{d^2 x^a_i(t)}{dt^2}  x^b_i(t)   \sqrt{g_i} \right \} + \int_0^T dt\;  \Delta S_{int}. \label{eq:GR0}
 \eea
Near equilibrium the total entropy production must vanish $\Delta S_x[g] \approx 0$ and then the entropy production due to interactions can be expressed as 
\be
\Delta S_{int} \approx  - \int d^3 x  \left \{ g_{i, ab}  \frac{d^2 x^a_i(t)}{dt^2} x^b_i(t)  \sqrt{g_i} \right \}=  - \int d^3 x  \left \{ g_{i, ab}  \frac{d^2 x^a_i(t)}{dt^2}  x^b   \sqrt{g_i} \right \}.\label{eq:Sint}
\ee
Using the geodesic equation \eqref{eq:geo} and \eqref{eq:Sint} the total entropy production \eqref{eq:GR0} can be recast into the following form,
 \bea
\Delta S_x[g]
 &\approx& -   \int_0^T  dt d^3 x  \left ( \Gamma^0_{\alpha\beta}(x) \left \{ \frac{d x_i^{a}}{dt} g_{i, ab}  \left ( x^b -  x^b_i(t) \right )   \frac{d x_i^{\alpha}(t)}{dt} \frac{d x_i^{\beta}(t)}{dt}  \sqrt{g_i} \right \}\right. \label{eq:GR1}\\
&&\;\;\;\;\;\;\;\;\;\;\;\;\;\;\;\;\;\;\;\;\;\;\;\;\;\;\; \left . - \Gamma^a_{\alpha\beta}(x) \left \{  g_{i, ab}  \left ( x^b -  x^b_i(t) \right )   \frac{d x_i^{\alpha}(t)}{dt} \frac{d x_i^{\beta}(t)}{dt}  \sqrt{g_i} \right \}  \right ). \notag
 \eea
 
To proceed further, we make a crucial assumption that on average,
 \be
 \frac{d x_i^{\alpha}(t)}{dt} \frac{d x_i^{\beta}(t)}{dt}  \approx \frac{1}{2} \left (g_{i}^{\alpha\beta}+ g^{\alpha\beta}(x_i) \right ). \label{eq:assumption}
 \ee
In other words, we assume that displacement of $i$'th neuron depends equally on its own covariance matrix $g_{i}^{\alpha\beta}$ and on the weighted average covariance matrix $g^{\alpha\beta}(x_i)$. Then by plugging \eqref{eq:assumption} into \eqref{eq:GR0} and using \eqref{eq:detg} and \eqref{eq:ginv} we get, 
 \bea
\Delta S_x[g] &\approx&  - \frac{1}{2} \int_0^T dt d^3 x \left (  g^{\alpha\beta}  \Gamma^0_{\alpha\beta} \left \{ \frac{d x_i^{a}}{dt} g_{i, ab}  \left ( x^b -  x^b_i(t) \right )    \sqrt{g_i} \right \} -g^{\alpha\beta}  \Gamma^a_{\alpha\beta} \left \{  g_{i, ab}  \left ( x^b -  x^b_i(t) \right )   \sqrt{g_i} \right \}  \right. \notag \\
&& + \left .  \Gamma^0_{\alpha\beta} \left \{ \frac{d x_i^{a}}{dt} g_{i, ab}  \left ( x^b -  x^b_i(t) \right )   g_{i}^{\alpha\beta} \sqrt{g_i} \right \} - \Gamma^a_{\alpha\beta} \left \{  g_{i, ab}  \left ( x^b -  x^b_i(t) \right )  g_{i}^{\alpha\beta}  \sqrt{g_i} \right \}  \right ) \notag \\
&=&  - \frac{1}{2} \int_0^T dt d^3 x \left (  g^{\alpha\beta} \Gamma^\mu_{\alpha\beta} \frac{\partial }{\partial x^\mu } \left \{  \sqrt{g_i} \right \} +     \Gamma^\mu_{\alpha\beta} \frac{\partial }{\partial x^\mu } \left \{ g^{\alpha\beta}_i \sqrt{g_i} \right \} \right ) \notag\\
 &=&  - \frac{1}{2} \int_0^T dt d^3 x \, \left (  \sqrt{-g} g^{\alpha\beta}   \Gamma^\mu_{\alpha\beta} \Gamma^\gamma_{\gamma\mu} +  \Gamma^\mu_{\alpha\beta}  \frac{\partial}{\partial x^\mu} \left ( \sqrt{-g} g^{\alpha\beta} \right ) \right ).  \label{eq:GR2}
 \eea
It is now easy to show that  \eqref{eq:GR2}  is equivalent to the Einstein-Hilbert action up to a boundary term, i.e.
 \bea
\Delta S_x[g] &\approx& -\frac{1}{2}\int dt d^3 x \, \left (  \sqrt{-g} g^{\alpha\beta}   \Gamma^\mu_{\alpha\beta} \Gamma^\gamma_{\gamma\mu} +   \Gamma^\mu_{\alpha\beta}  \frac{\partial}{\partial x^\mu} \left ( \sqrt{-g} g^{\alpha\beta} \right ) \right ) \label{eq:GR3}\\
&=&  \int dt d^3 x \, \sqrt{-g} g^{\alpha\beta} \left (  \Gamma^\mu_{\alpha\beta} \Gamma^\gamma_{\gamma \mu} -  \Gamma^\mu_{\alpha\gamma} \Gamma^\gamma_{\beta \mu}  + \frac{\partial}{\partial x^\mu} \Gamma^\mu_{\alpha\beta}  - \frac{\partial}{\partial x^\beta}  \Gamma^\mu_{\alpha\mu}   \right ) + \text{boundary term}\notag
\\
&=& \int dt d^3 x \, \sqrt{-g} g^{\alpha\beta} R_{\alpha\beta} + \text{boundary term} \notag
\eea
where the Ricci tensor is 
\be
R_{\mu\nu} \equiv  \Gamma^\mu_{\alpha\beta} \Gamma^\gamma_{\gamma \mu} -  \Gamma^\mu_{\alpha\gamma} \Gamma^\gamma_{\beta \mu}  + \frac{\partial}{\partial x^\mu} \Gamma^\mu_{\alpha\beta}  - \frac{\partial}{\partial x^\beta}  \Gamma^\mu_{\alpha\mu}.
\ee
By setting variations of the entropy production $\Delta S_x[g]$ with respect to the inverse metric tensor $g^{\mu\nu}$ to zero we obtain the vacuum Einstein equations,
\be
\frac{\delta \Delta S_x[g] }{\delta g^{\mu\nu}} = 0 \;\;\;\;\;\;\;\;\;\;\Rightarrow\;\;\;\;\;\;\;\;\;\;\;\;\;R_{\mu\nu} - \frac{1}{2} g^{\alpha\beta} R_{\alpha\beta} g_{\mu\nu} = 0. \label{eq:einstein}
\ee
(See Ref. \cite{Carroll} for a detailed derivation of the Einstein equations from the Einstein-Hilbert action.) 

So far the total number of neurons $N$ was fixed, but, as was argued in Ref. \cite{quantumness} and in Sec.  \ref{sec:schrodinger}, for the quantumness to emerge the number of neurons $N$ must vary.  Such variations can be introduced into the variational problem by defining a functional,
\bea
{\cal S}[g, \Lambda] &\equiv& \Delta S_x[g]  +  2 \Lambda \left ( \bar{N} - N\right )\notag\\
& =&  \int dt d^3 x \;  \sqrt{-g} \left ( g^{\alpha\beta} R_{\alpha\beta} - 2 \Lambda \right ) +  2 \Lambda \bar{N},
\eea
where we used \eqref{eq:GR3}, \eqref{eq:N} and \eqref{eq:detg}. By varying ${\cal S}[g, \Lambda]$ with resect to the inverse metric $g^{\mu\nu}$, we obtain Einstein equations with cosmological constant, i.e.
\be
R_{\mu\nu} - \frac{1}{2} g^{\alpha\beta} R_{\alpha\beta} g_{\mu\nu} + \Lambda g_{\mu\nu} = 0.
\ee
Evidently, the Lagrange multiplier $2 \Lambda$ constraints the average number of neurons and plays a role of the cosmological constant $\Lambda$ in the gravitational description of non-trainable variables. We recall that in the quantum description of trainable variables (see Sec. \ref{sec:schrodinger}) the Lagrange multiplier $\mu = \pm 2\pi \hbar$ also constraints the average number of neurons, but instead it plays the role of the Planck's constant $\hbar$.  Evidently, the role of the Lagrange multipliers $2 \Lambda$ and $ \mu = \pm 2\pi \hbar$ in the gravitational description of non-trainable variables and in quantum description of trainable variables is very different. 

In  statistical description, the parameter $\Lambda$ would play the role of a ``chemical potential'' which would be responsible for both ``neurogenesis'' and ``neurodegeneration''. If the parameter can vary in time, then for a system with a small number of neurons (e.g. early Universe) $\Lambda$ would be larger, but for a systems with a large number of neurons (e.g. late Universe) $\Lambda$ would be smaller. This can potentially explain both: the early-time accelerated expansion (i.e. cosmic inflation) and the late-time accelerated expansion (i.e. the dark energy), but for the former case a more thorough modeling of the spatial variations of the number density of neurons is required.  In addition, the dynamics of trainable variables ${\bf q}(t)$ must be described by either Madelung or Schrodinger equations (see Secs. \ref{sec:madelung} and \ref{sec:schrodinger})  and thus additional equations of motions must be satisfied and additional constraints must be imposed.  However, from the point of view of  the metric dynamics, there should exist an appropriately defined energy momentum tensor $T_{\mu\nu}$ that would be acting as a source in the Einstein equations,
\be
R_{\mu\nu} - \frac{1}{2} g^{\alpha\beta} R_{\alpha\beta} g_{\mu\nu}+ \Lambda g_{\mu\nu}  = \kappa T_{\mu\nu}.
\ee
In addition, it is important to model possible deviations from the assumption \eqref{eq:assumption} in the context of astrophysical observations of, for example, dark matter.  Of course, all such generalizations require a more careful modeling of the dynamics of the trainable variables which is beyond the scope of this paper. 

\section{Discussion}\label{sec:discussion}

All successful physical models are built on top of mathematical frameworks or theories. These theories are never proven in a rigorous mathematical sense, but instead they are validated through either repeated experiments or observations of the Universe around us.  In the twenties century two such theories were first proposed - {\it quantum mechanics} and {\it general relativity} - and then successfully applied to modeling physical phenomena on a wide range of scales from $10^{-19}$ meters (i.e. high-energy experiments) to $10^{+26}$ meters (i.e. cosmological observations). However, all of the attempts to treat one of these theories as fundamental, and the other one as emergent have so far failed (i.e. the problem of quantum gravity). In addition, both theories seem to fall apart with introduction of macroscopic observers like ourselves. In some sense, the situation with observers was even worse than with physical phenomena, since we did not even have a mathematical framework for modeling observers. Indeed, there is not a single self-consistent and paradox-free definition of macroscopic observers that could describe what is actually happening with quantum state during measurement (i.e. the measurement problem) or how to assign probabilities to cosmological observations (i.e. the measure problem). Fortunately, the situation is changing and now we do have a mathematical framework of {\it neural networks} which can describe many (if not all) biological phenomena \cite{evolution}. The main question, however, remains: can the theory of neural networks be the fundamental theory \cite{worldasnn} from which (not only macroscopic observers \cite{originoflife} or some complex phenomena \cite{criticality}, but) {\it all} biological and physical phenomena emerge? If so, then the theories of quantum mechanics and general relativity must not be fundamental, but emergent. 
  
The idea that quantum mechanics can emerge from anything classical, including neural networks, is very counterintuitive. And the main problem is not that in quantum mechanics we are dealing with probabilities and in classical physics everything is deterministic. Even in quantum mechanics the wave-function $\Psi({\bf q})$  evolves deterministically and it is only because of the measurements  the probabilities $p({\bf q})=|\Psi({\bf q})|^2$ arise. In fact, this is not very different from statistical mechanics, but what is difference is that in quantum mechanics not only probabilities, or square-root of probabilities $|\Psi({\bf q})|$, but also the complex phase of the wave-function $\text{Im} (\log (\Psi({\bf q})))$,  evolves according to the Schrodinger equations. To show that this might be possible in a given dynamical system requires two non-trivial steps. The first step is to provide a microscopic interpretation of the complex phase which, in the case of neural networks, is the free energy of non-trainable variables $\text{Im} (\log (\Psi({\bf q}))) = \tilde{F}({\bf q})/\hbar$. Note that the microscopic interpretation of the phase was also given in Ref. \cite{entropic} for constrained systems and  in Ref. \cite{learningtheory, worldasnn} for equilibrium systems, but as was shown in Sec. \ref{sec:madelung} similar results also hold for non-equilibrium systems. The second step is to show that the complex phase, or the free energy $\tilde{F}({\bf q})$ in the case of neural networks, is multivalued. The multivaluedness condition is essential for the fully quantum behaivor to emerge \cite{wallstrom} and in the case of neural networks it is satisfied for a grand-canonical ensemble of neurons \cite{quantumness}. In Sec. \ref{sec:schrodinger} we extended this result to non-equilibrium systems that are capable of adjusting its own parameters (e.g. number of neurons, step size, mini-batch size). More precisely, we have shown that the quantum description of neural networks is appropriate for modeling the non-equilibrium dynamics of trainable variables with non-trainable (or hidden) variables modeled through their free energy and the number of neurons constrained by a Lagrange multiplier which plays the role of the Planck constant. 

The problem of emergent gravity \cite{Jacobson, Verlinde, Padmanabhan, emergent} is even more complicated, just because it is impossible to study the emergence of general relativity until the space, time and space-time symmetries had already emerged. In the context of neural networks, the problem  was first studied in Ref. \cite{learningtheory} and more specifically in Ref. \cite{worldasnn}, but in both cases the description was too phenomenological or architecture-dependent for anything substantial to be said about the nature of dark energy, dark matter or cosmic inflation. In this paper we improved our understanding of the emergent gravity on several fronts. First of all, we showed that the Lorentz symmetries emerge from the equilibrium dynamics for null vectors, from the stochastic entropy production for time-like vectors and from the entropy destruction due to learning for space-like vectors. This is in agreement with a common view that ``time'' has a thermodynamic origin, but it also suggests that ``space'' must emerge from learning. Secondly, we used the freedom of Lorentz transformations to define the emergent space-time and the metric tensor which is, by construction, as close as possible to being flat. In fact, it was essential for the space-time to be nearly flat and we expect the relativistic description to break down in regions of high curvature. Thirdly, we considered localized states of neurons, with minimal interactions described by the metric tensor, to show that they must move along geodesics in the emergent space-time. And finally, we showed that the general relativistic description is appropriate for modeling the dynamics of non-trainable variables with trainable variables modeled through their energy-momentum tensor and with the number of neurons constrained by a Lagrange multiplier which plays the role of the cosmological constant. 

In conclusion, we would like to emphasize that the quantum and gravitational descriptions presented in this paper are dual in the sense that they provide alternative macroscopic descriptions of the same learning system, defined microscopically as a neural network. This duality does not have an obvious connection to the holographic duality \cite{Maldacena, Witten, Susskind} although such possibility was discussed in Ref. \cite{worldasnn}. On the other hand, a fully quantum descriptions can only emerge from a neural network if the number of neurons is not fixed in which case a constraint on the number of neurons must be imposed in both sectors, i.e. gravitational and quantum. The Lagrange multiplier which imposes the constraint in the quantum description is the Planck constant (see Sec. \ref{sec:schrodinger}), but the Lagrange multiplier which imposes the constant in the gravitational description is the cosmological constant (see Sec. \ref{sec:einstein}). This implies that a quantum system can only be dual to a gravitational system with cosmological constant as in AdS/CFT \cite{Maldacena, Witten, Susskind}, but the sign of the cosmological constant can be arbitrary.
\\
\\
{\it Acknowledgments.}  The author is grateful to Mikhail Katsnelson and Kirill Zatrimaylov for very useful discussions. This work was supported in part by the Foundational Questions Institute (FQXi) and the Oak Ridge Institute for Science and Education (ORISE).

\end{document}